%% file: coling_latex.tex
\pdfoutput=1

\documentclass[11pt]{article}

\usepackage{coling}

\usepackage{times}
\usepackage{latexsym}
\usepackage{listings}
\usepackage{cleveref}
\usepackage{multirow}

\usepackage[T1]{fontenc}

\usepackage[utf8]{inputenc}

\usepackage{microtype}

\usepackage{inconsolata}

\usepackage{graphicx}

\title{ALoFTRAG: Automatic Local Fine Tuning for Retrieval Augmented Generation}

\author{Peter Devine \\
  Lightblue KK. \\
  Tokyo \\
  \texttt{peter@lightblue-tech.com}
  }

\begin{document}
\maketitle
\begin{abstract}
Retrieval Augmented Generation (RAG) systems have been shown to improve the accuracy of Large Language Model (LLM) outputs. However, these models can often achieve low accuracy when applied to new data domains.

We introduce the Automatic Local Fine Tuning of Retrieval Augmented Generation models (ALoFTRAG) framework, designed to improve the accuracy of RAG systems on a given domain by training LLMs without manually labeled data or using larger teacher models. 

By generating and filtering synthetic training data and performing LoRA fine-tuning, ALoFTRAG improves citation and answer accuracy across 20 datasets in 26 languages by, on average, 8.3\% and 3.0\% respectively. 

Our results demonstrate that ALoFTRAG offers a practical, cost-effective, and data-secure solution for improving RAG accuracy, making it particularly applicable to sensitive domains such as healthcare and finance.
\end{abstract}

\section{Introduction}

Retrieval augmented generation (RAG) models are a subset of large language models (LLMs) which combine the generation capabilities of conventional LLMs with the factual grounding of information retrieval (IR) models to create more factually accurate outputs from LLMs~\cite{lewis2020retrieval}. RAG models work by taking a user question as input, and then selecting several reference texts with high semantic similarity (determined by an IR model) from a database. An LLM is then given these texts with the original question and is instructed to answer the question basing the answer on the relevant reference texts.

RAG not only allows for more accurate answers to questions regarding general public knowledge~\cite{guu2020retrieval,ram2023context}, it also allows LLMs to generate responses based on locally available or domain specific information that it has not necessarily been trained upon~\cite{gao2023retrieval,zhang2024raft}.

However, the models that have exhibited the highest performance in RAG tasks are based on proprietary cloud-based LLMs, meaning that LLMs run locally are more likely to generate hallucinations or other untruthful outputs when being used for RAG~\cite{HughesBae2023}. Moreover, LLMs that are not trained using data from a specific domain exhibit lower RAG accuracy in that domain~\cite{zhang2024raft}.

To address this, we propose a framework called Automatic Local Fine Tuning of Retrieval Augmented Generation models (ALoFTRAG).
ALoFTRAG improves the accuracy of base RAG systems by automatically training on the data which the system will later be used, all without using larger models or labelled data. 

We demonstrate the effectiveness of ALoFTRAG by performing experiments on 20 datasets in 26 languages across a variety of domains and comparing the accuracy to simply using the base LLM for RAG. We show that the ALoFTRAG approach improves both the citation accuracy and answer accuracy of RAG models across almost all datasets compared to the base RAG model.

Our findings inform the future implementation of RAG systems, allowing users to fine-tune their RAG models on local data using modest hardware, enabling improved RAG accuracy while preserving data security.

\begin{figure*}[t]
\centering
  \includegraphics[width=1.0\linewidth]{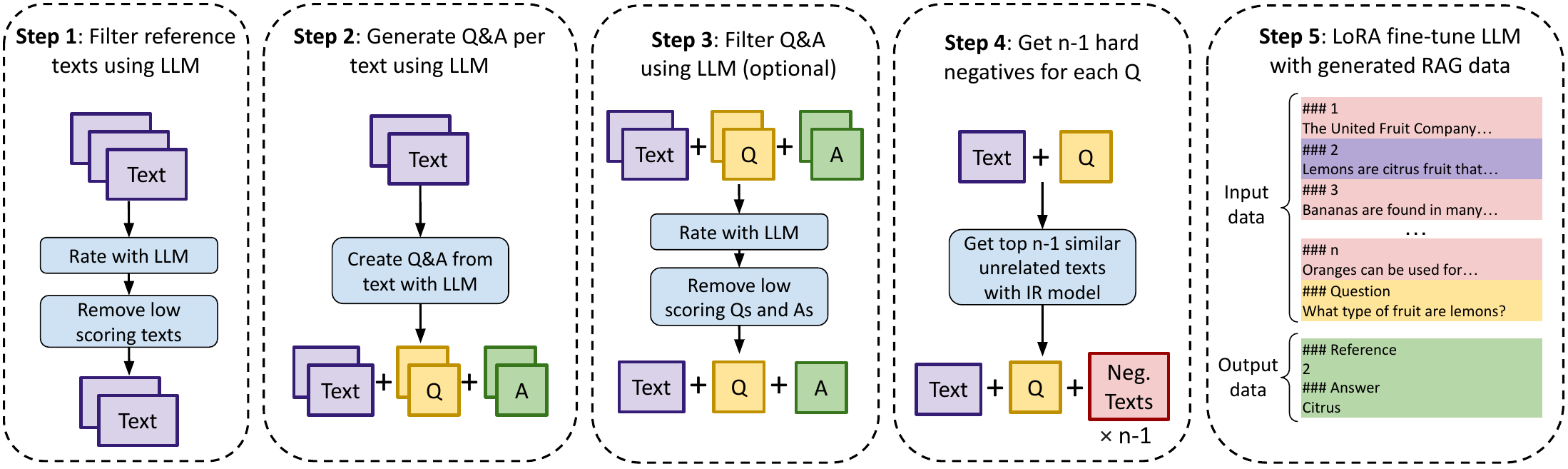}
  \caption {An illustration of the ALoFTRAG framework.}
  \label{fig:aloftrag_diagram}
\end{figure*}

\section{Related work}

RAG was first proposed as a technique to improve the output of LLMs to users~\cite{lewis2020retrieval}. This technique has been shown to reduce hallucinations of models and thus increase the veracity of outputs in conversation~\cite{shuster2021retrieval}.

Benchmarks have shown that proprietary remote models (i.e. models running on cloud servers) such as GPT-4 Turbo consistently outperform open source local models (i.e. models that can be run on consumer grade compute) such as Llama 3 70B Instruct~\cite{yang2024crag}. Previous work has shown that training a local RAG model on a specific domain can improve the accuracy of that model on the domain~\cite{siriwardhana2023improving}.

One way to obtain domain-specific data is to generate it using LLMs. Work such as Self-Instruction has been shown to improve the chat abilities of an LLM by training on filtered synthetic data from the same model~\cite{wang2022self}.

Previous work has shown that generated synthetic RAG data can be used for the purposes of evaluating RAG systems on specific domains~\cite{zhu2024rageval}. Other work has demonstrated that it is possible to preserve privacy by anonymizing the input data of an RAG-enabled health chat system using local LLMs which can then be safely uploaded to proprietary remote LLMs~\cite{zeng2024mitigating}. However, these approach does not seek to improve the accuracy the actual RAG system.

We show that RAG accuracy can be improved without manual labeling or proprietary models by generating data from unlabelled text using a local LLM, which then trains itself on that data. 
This increases the model's accuracy on the text's domain while ensuring privacy by keeping both training and inference on local hardware.

\section{ALoFTRAG}

In this work, we propose the Automatic Local Fine Tuning of Retrieval Augmented Generation models (ALoFTRAG) framework, which we designed to increase the accuracy of LLM-based RAG systems while using only one locally available base LLM and an IR model.
This section details the process involved in performing ALoFTRAG and summarizes the reasons for carrying out each step.

The ALoFTRAG process starts with a set of unlabelled that will be used as the reference texts in a RAG system. We then apply 5 steps to prepare training data: Filtering reference texts, generating Q\&As, filtering Q\&As, selecting hard negatives, and fine-tuning as RAG. We detail these steps below:

\subsection{Step 1: Filtering reference texts}

Before starting the ALoFTRAG process, we assume that the reference text documents have been chunked into tractable sizes for an LLM.

We start the ALoFTRAG process by providing the prompt described in \cref{lst:textfiltersysmsg} as the system message for a base LLM, which instructs the LLM to generate a rating between 1-10 for a given piece of text depending on how much useful information it contains. We generate a rating for each reference text using this LLM with the vLLM inference package\footnote{\url{https://github.com/vllm-project/vllm}}~\cite{kwon2023efficient} with generation temperature set to zero, which we used for generation tasks throughout our experiments. We then parse the numerical rating and filter any ratings that fall below a certain threshold. We also remove any instances where we fail to parse a rating given the output.

In our experiments, we set this threshold at 8, as we found it to be low enough that some datasets had large amount filtered out (57\% filtered out on the CaLMQA Chinese subset) while being high enough that most datasets kept the vast majority of their data (we average only 6.9\% of texts filtered out across all data subsets). We perform ablation tests in our experiments to determine the effect of this step.

\subsection{Step 2: Generating Q\&As}

We then set the base LLM system message to that prompt described in \cref{lst:qagensysmsg}, with the \texttt{\{language\_name\}} replaced with the name of the dataset language in English. This instructs the model to write a self-contained question and answer that can be asked and answered purely by reading the reference text. We generate a question and answer for each reference text and discard that were not correctly parsed in the format requested in the system message. This lead to a maximum of 33\% questions not being parsed at this stage for the CaLMQA Kirundi subset, with an average of 1.1\% over all data subsets.

With this, we have a set of reference text, question, answer triplets for each dataset.

\subsection{Step 3: Filtering Q\&As}

To filter the generated questions and answers, we set the system message to those described in \cref{lst:qratingsysmsg} and then \cref{lst:aratingsysmsg}. This instructs the model to generate a rating between 0-10 for each question and answer independently. The question rating is instructed to be based on the answerability and fluency of the question. The answer rating is instructed to be based on the veracity and fluency of the answer. 

We set our rating threshold for both question and answer ratings to 8 and filter out any instances where either the question or answer fall below this threshold. A mean of 2.4\% of questions and 1.4\% answers were filtered out across all data subsets (maximums of 11.2\% and 7.0\%, respectively). We also perform ablation tests in our experiments to determine the effect of this step, where we find that it is detrimental to reference and answer accuracy in a majority of cases. For this reason, we consider this step optional when performing ALoFTRAG.
\input{table_dataset}


\subsection{Step 4: Selecting hard negatives}

With a filtered list of reference text, question, answer triples, we sampled multiple hard negatives for each question. This was done by embedding each reference text and embedding each generated question in our dataset using a dense text embedding model. We then obtained the similarity between each question and reference text by the matrix product of the respective embeddings. This gave us a list of similar reference texts for each generated question, from which we removed the correct reference text that was actually associated with the generated question. We sample the $n-1$ most similar reference texts from this list as our hard negatives and we add the positive reference text to make $n$ total reference texts given as context to the RAG model. We set $n=10$ contexts for most of our experiments as this is the largest number of reference texts we could viably use across all datasets without exceeding maximum token memory constraints for the LLM. We vary this value to $n=5$ and $n=2$ contexts in ablation tests.


\subsection{Step 5: Fine-tuning for RAG}
\label{sec:ragfinetuning}

We use our generated questions, answers, reference texts and hard negative texts as training data for performing RAG. We prepare the training data by randomly shuffling the correct reference text into the hard negative texts and noting the index of the correct reference. We then format the training data as a conversation between a user and an assistant.

We first set the system message to be that described in \cref{lst:ragsysmsg}, which is a general RAG-style system message. Although fine-tuning should eliminate the requirement for using a system message to explain the task at hand, we include the system message in our training as it allows for a more direct comparison between the trained model and the base model during our evaluation.

After the system message, we input the user message data. The user message data is a Markdown-styled enumerated list of reference texts separated by newlines, followed by the question given, as illustrated in Step 5 of \cref{fig:aloftrag_diagram}.
Finally, we format the model response as the ordinal of the correct reference text and the generated answer for the question given, again styled like Markdown as shown in Step 5 of \cref{fig:aloftrag_diagram}.

We use this training data to then LoRA fine-tune~\cite{hu2021lora} an instruction-trained LLM using the same chat template it was originally trained on. By fine-tuning in this way, we train a model to answer a question by finding the correct reference text from a list of plausible candidates and then generate a correct answer to that question. We believe that training the model to first cite the correct reference text before answering adds explainability to the RAG system and may benefit training by adding a curriculum learning~\cite{bengio2009curriculum} approach to RAG training.

We perform the LoRA training in our experiments for 1 epoch using the Axolotl training framework\footnote{\url{https://github.com/axolotl-ai-cloud/axolotl}}. Details of our training parameters can be found in \cref{sec:trainingparams}.

\section{Evaluation}

We carry out our ALoFTRAG process on 20 datasets in 26 languages and evaluate the accuracy of the resultant models on the original gold-label questions and answers from each dataset. We compare these accuracies to that of the base LLM and several ablation tests varying different aspects of the framework.

\subsection{Data generation models}
The base LLM that we used for our experiments was the 7 billion parameter instruction tuned Qwen 2 model~\cite{yang2024qwen2}\footnote{Sourced from \url{https://huggingface.co/Qwen/Qwen2-7B-Instruct}}. We chose to use this model as it is multilingual, highly performant for its size across many tasks and benchmarks, small enough to run on a single consumer-grade GPU, and has a permissive Apache 2 which allows for commercial applications of the model. Since we fine-tune on top of this model, we refer to the instruction tuned Qwen 2 model as the ``base model'' throughout this paper.

The IR model we used for step 4 of our ALoFTRAG process was the dense BGE-M3 embedding model~\cite{bge-m3}\footnote{Sourced from \url{https://huggingface.co/BAAI/bge-m3}} which we chose because it is multilingual, has high evaluation scores across a variety of benchmarks, and is small enough to run on a consumer grade GPU at the same time as the above 7 billion parameter LLM.
While alternative IR models could be used in place of a dense embedding model, we leave it to future work to improve upon the IR aspect of the ALoFTRAG implementation.

\subsection{Datasets}

To evaluate the ALoFTRAG approach, we trained and tested on 20 popular question answering datasets in 26 unique languages. 
From each dataset, we extract a gold-label question, gold-label answer, and reference text for each row in the dataset. 
We selected all of our datasets for their diversity in languages and content domains.

Extra details of our dataset pre-processing can be found in \cref{sec:datasetspecificpreprocessing}.

By order of preference, we select one of the test, validation, or train set for each available dataset.

We perform ALoFTRAG on each monolingual subset of each dataset using only the reference texts of each subset as positive and negative contexts. This resulted in a trained ALoFTRAG model for each language in each dataset.
We evaluate upon each language subset of every dataset separately, using the gold-label questions, gold-label answers, and reference texts for each.

Details of the datasets used in our evaluation can be found in \cref{tab:datasets}.

\subsection{Evaluation method}
\input{table_acc}

To evaluate our approach, we performed RAG using both the ALoFTRAG model and the base RAG model using largely the same RAG configuration as described in the ALoFTRAG training step. We first input the same system message as described in \cref{sec:ragfinetuning}. We then sample the 10 most similar reference texts to each gold-label question in the dataset. 

In situations where the 10 most similar contexts exceeds the maximum memory size of our model minus a margin for questions and answers (20,000 maximum tokens minus 1,000 token margin), we take the maximum number of contexts that do not exceed this limit, from most similar to least. 

In situations where the correct reference text is not within the most similar 10 texts, we swap the least similar text with the correct reference text. We term these cases as ``hard'' questions and contrast them with cases where the correct reference text is within the top 10 most similar texts as ``easy'' questions in our results. This distinguishes between questions which have high semantic similarity to the correct reference text and those that do not.

We randomly shuffle the contexts to remove any positional bias for referencing the contexts and and input them into the LLM in an enumerated Markdown-header-styled list with the question.

\input{table_hard_easy_acc}

As with ALoFTRAG, we use the vLLM inference package to generate responses from these inputs, generating responses using either the base LLM and the LoRA  trained model. This gave us both reference citations and textual answers to each gold-label question for both the base LLM and the ALoFTRAG LLM model.

We evaluated over each language subset in each dataset, then average the scores of each language subset to create a dataset score.

We evaluate the \textbf{reference accuracy} by calculating the percentage of instances where the correct reference text ordinal is contained within the reference ordinals output by the model.

We evaluate the \textbf{answer accuracy} by providing the gold label reference text, question and answer, as well as the generated answer to the \verb|gpt4o-2024-05-13| version of GPT4o~\cite{openai-2024}. We set the system message of GPT4o to that described in \cref{lst:anschecksysmsg} and we calculate the percentage of GPT4o responses that judge the generated response as correct.

\section{Results}

The reference and answer accuracies for each dataset can be found in \cref{tab:acc_results_dataset_averaged}. This includes the accuracies of the base model, the full ALoFTRAG models, as well as the ALoFTRAG ablation tests where either the text filtering (Step 1) or question and answer filtering (Step 3) steps are removed from our ALoFTRAG process. The full results for each language subset can be found in the appendix in \cref{tab:fullresultsappendix} and \cref{tab:fullresultsappendix2}.

We find that every ALoFTRAG implementation achieves higher citation accuracy and answer accuracy across almost all datasets compared to the base model.

We observe that when the text filtering step (Step 1) is removed from the ALoFTRAG process, answer accuracy reduces and citation accuracy increases, on average.
We also observe that when the question and answer filtering step (Step 3) is removed from the ALoFTRAG process, answer accuracy and citation accuracy both increase on average.

Paired t-tests~\cite{Ross2017} show that the differences of the mean value of the all step ALoFTRAG model to the base model, the model without step 1, and the model without step 3 are statistically significant to $p<0.05$.

The average differences between hard and easy accuracies for both answer and reference accuracies can be found in \cref{tab:hard_easy_results}. We demonstrate that easy questions unsurprisingly have a higher reference and answer accuracy compared to hard questions. We also show that the jump in answer accuracy when performing ALoFTRAG across all models is more pronounced for hard questions than easy questions. Conversely, we find that the jump in reference accuracy between base and ALoFTRAG models is lower for hard questions than easy questions.

Our results also show that there are many cases in which the model achieves a higher answer accuracy than reference accuracy. While we initially thought that this may be due to the model simply referencing a text that contained the answer to the question but was not the selected reference text. However, we found from analysis of the JSQuAD results that the model was referring to texts that did not contain the correct answer, but then outputting the correct answer anyway. Averaged over datasets, we find that the base model referenced the wrong text but gave the right answer in 19.7\% of cases, while this occurred in 10.1\%, 8.9\%, and 6.3\% of cases for ALoFTRAG all steps, without step 1, and without step 3, respectively.

\begin{figure*}
  \centering
  \includegraphics[width=.45\linewidth]{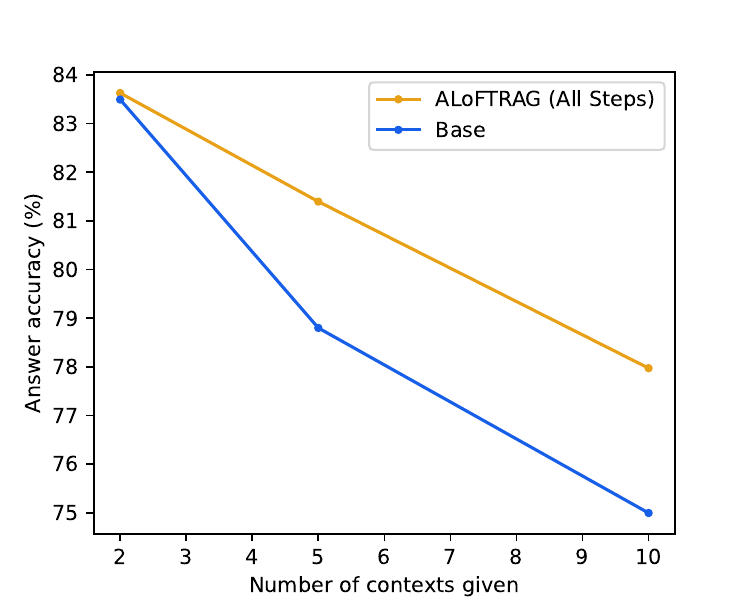}
  \centering
  \includegraphics[width=.45\linewidth]{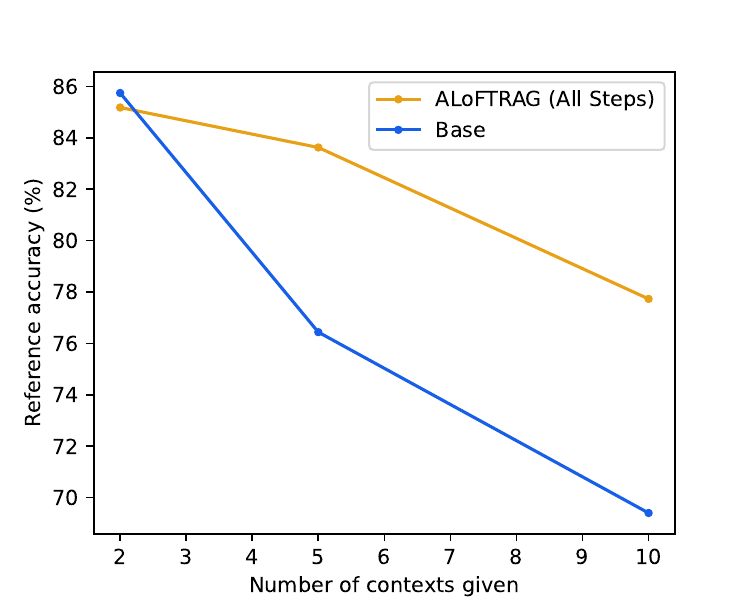}

\caption{Plots of answer and reference accuracy varied over number of chunks. Note that the correct context was always within the contexts, making the 2 context task necessarily simpler than the 10 context task.}
\label{fig:chunk_num_diagram}
\end{figure*}

We can see from \cref{fig:chunk_num_diagram} that as we introduce more distractors (i.e. incorrect reference texts) to the reference texts given to the RAG system, the reference and answer accuracy decrease for both the base model and the all steps ALoFTRAG model. This confirms previous work where moving from 2 to 10 contexts resulted in lower RAG accuracy~\cite{fatehkia2024t}.
What is notable is that when 2 contexts are given, effectively meaning that there is only 1 distractor context and 1 correct context within the given contexts, the accuracy of the base model is roughly the same as or higher than the ALoFTRAG model. However, in more difficult RAG configurations when there are 5 or 10 given contexts, the accuracy of the ALoFTRAG model is noticeably higher than the base model.

\section{Discussion}
Our first finding is that performing ALoFTRAG improves both the citation accuracy and answer accuracy of the base LLM in nearly all cases. This indicates the utility of ALoFTRAG to RAG practitioners as we have demonstrated that it makes the output of the final model more accurate.

Our results also show that excluding the question and answer filtering step (step 3) leads to higher performance on both answer and reference accuracy. While this step aimed to filter out incorrect or noisy data, our current implementation seems to remove beneficial data. Since previous work has shown that LLMs can be improved using their own outputs~\cite{zelikman2022star,li2023self,pang2023language,wang2022self}, we aim to refine this step in future work, but we have set our released ALoFTRAG code to not perform step 3 by default.

A curious finding is that excluding step 1, filtering out noisy text, leads to lower answer accuracy but higher reference accuracy. This suggests that noisy training data may benefit in choosing the correct reference but be detrimental in generating the correct answer.

Another finding is that the hard questions have lower answer accuracy and reference accuracy across all models. In our evaluation, the correct reference text is supplied to even when answering hard questions, meaning that the model output the wrong answer despite having the relevant context. Therefore, even in an environment with an extremely accurate IR model in which the correct text is always selected for RAG, the LLM answer generation will be a limiting factor in RAG accuracy, highlighting the importance of improving answer accuracy with ALoFTRAG to RAG practitioners.

From our results, we found that ALoFTRAG training reduces the amount of times where the model references the wrong text but gives the correct answer. This gives the RAG system more accountability, indicating that ALoFTRAG can help the user to know exactly \textit{how} the model has come to the answer it outputs.

We also find that as the number of chunks in the RAG setting increases, the difference in answer and reference accuracy between ALoFTRAG and base models grows larger. This suggests that ALoFTRAG's improvements are partly due to training with many distractors. Previous RAG systems can contain up to 100 chunks~\cite{finardi2024chronicles}, so ALoFTRAG's effect may be heightened with larger numbers of contexts.

Overall, ALoFTRAG training generally improves both answer and reference accuracy of RAG systems. This is significant as our approach does not rely on costly manually labeled data, larger LLMs for model distillation~\cite{hinton2015distilling}, and is general enough for various use-cases and languages. It is thus accessible and cost-effective for a wide range of RAG users.


Moreover, LoRA fine-tuning in ALoFTRAG requires much less memory than full fine-tuning~\cite{hu2021lora}\footnote{See \url{https://github.com/hiyouga/LLaMA-Factory\#hardware-requirement}}, enabling RAG practitioners to use consumer-grade GPUs. This makes ALoFTRAG accessible and data secure for even the "GPU poor."

The security aspects of our approach are also significant as many individuals and organisations perform RAG on data that contains sensitive information, including medical records, financial data, and confidential commercial information. It is thus inappropriate to train an RAG model on data in this domain by augmenting the data using a remote, proprietary model. However, ALoFTRAG can be trained in a completely closed loop, meaning that ALoFTRAG can improve answer accuracy without compromising data security.

We consider a few investigations for future work, including whether this approach could scale to training IR models.
Our results have shown that using our correct context, generated questions, and generated answer triplets to train the LLM part of the RAG system improves LLM answer and reference accuracy. Future work could examine using this data to also train the IR model aspect of the system to improve context retrieval accuracy.

Realistic RAG would also include instances where the correct answer cannot be given as the correct reference text has not been supplied. We leave it to future work to add such instances into the RAG training framework.

Future work could also examine the applicability of an approach similar to ALoFTRAG for multimodal RAG~\cite{chen2022murag} by possibly generating questions and answers given images, videos, or tabular data and training on this data.

Finally, future work could perform ALoFTRAG using a large amount of synthetically generated open source RAG data, resulting in a strong generalist RAG model which could then be further fine-tuned with local ALoFTRAG on potentially sensitive data.

\section{Conclusion}

In this work, we introduced the Automatic Local Fine Tuning of Retrieval Augmented Generation models (ALoFTRAG) framework, designed to improve the accuracy of RAG systems. 
Our experiments across 20 datasets in 26 languages demonstrated that ALoFTRAG consistently improves both citation and answer accuracy compared to the base LLM. 

By leveraging self-generated training data and performing LoRA fine-tuning, ALoFTRAG provides a practical, data-secure, and cost-effective solution for RAG practitioners. 
This framework is particularly beneficial for applications requiring data privacy, such as medical and financial domains. 
Our findings suggest that ALoFTRAG can democratize access to high-performing RAG systems, enabling more accurate and reliable outputs across a wide range of use cases and languages. 
Future work will focus on refining the ALoFTRAG process and exploring its application to multimodal use cases.

We make the code for reproducing our experiments and performing ALoFTRAG and the results of our experiments freely available online\footnote{\url{https://github.com/lightblue-tech/aloftrag}}.



\section{Limitations}
This section lists the current limitations of our work.

Firstly, the results of our work only show a few percent increase in citation and answer accuracy in most cases compared to the base LLM. Although this is notable that there is an increase over such a broad range of datasets and languages, it limits the impact of our work.

The ALoFTRAG approach is also rather naive in several respects, by generating only one question and answer per unique context, only training for one epoch, keeping a set number of reference texts, and having a fixed filtering threshold. There may be cases in which these parameters may be sub-optimal for performance, meaning that we could achieve higher accuracy with more optimal configurations, but we leave it to future work to investigate more sophisticated strategies for performing ALoFTRAG.

By their nature, public datasets are somewhat clean, and so we have not been able to test ALoFTRAG on potentially noisy proprietary data that many RAG practitioners would realistically use. We leave it to future work to evaluate upon additional datasets.

We attempted to apply ALoFTRAG to datasets in low resource languages with unique scripts, such as the Amharic AmQA~\cite{taffa2024low}. However, we found that our Qwen model could not reliably generate questions and answers in this script, so this ALoFTRAG would require some adjustment to the generation prompt or the use of a different generation model to be used on low resource languages.

\bibliography{custom}

\appendix

\section{Dataset specific pre-processing}
\label{sec:datasetspecificpreprocessing}

The CalmQA and Public Health QA datasets did not have reference texts associated with their questions, so we used the answers provided as both the reference texts and the answers due to the fact that the answers were generally long enough to be usable as an RAG reference text.

The KenSwQuad and chaii-1 datasets have a small minority of very long single reference texts (greater than 1,500 tokens) which we filtered out to avoid memory issues.

Many of the datasets include a title of the article from which the reference text has been taken, so where possible we include the title at the start of the reference text.

The Narrative QA dataset was included in our experiments as its questions are about non-general information specific to its fictional reference texts, meaning that a model needs to comprehend the reference text when answering rather than simply memorising large amounts of general knowledge.

CalmQA and M2QA were included as they were published at roughly the same time or after the release of the Qwen 2 model, meaning that it is highly unlikely that the Qwen 2 model we used was trained using these datasets.

\section{System messages}
\label{sec:sysmsgappendix}

\lstset{
  basicstyle=\small,
  columns=fullflexible,
  frame=single,
  breaklines=true,
  postbreak=\mbox{$\hookrightarrow$\space},
}

\begin{lstlisting}[caption={Text rating system message},captionpos=b,label={lst:textfiltersysmsg}]
You are a text filtering AI model.
Your input is a piece of text.
You output is a score of how much useful information is included within the text.

Output your score on a scale of 0-10, with 0 meaning that the text contains no useful information and 10 meaning that the text contains a large amount of useful information.

Your output should be formatted like so:

### Filter score
[YOUR SCORE]
\end{lstlisting}

\begin{lstlisting}[caption={Question and answer generating system message},captionpos=b,label={lst:qagensysmsg}]
You are a QA generating AI model.
Your input is a piece of text.
You output a question that can be answered solely by reading the text and the correct answer to that question.

Write the prompt so it does not refer to any knowledge that is assumed from the article.
Write the prompt so that it could be given without ever having read the passage.
Do not refer to the text directly (e.g. "According to the text", "Based on this passage").

If a short answer will suffice, then write a short answer.
Only write a long answer if required.

Your question and answer must be in fluent, natural {language_name}.

Your output should be formatted like so:

### Question
[YOUR QUESTION]

### Answer
[YOUR ANSWER]
\end{lstlisting}

\begin{lstlisting}[caption={Question rating system message},captionpos=b,label={lst:qratingsysmsg}]
You are a question and answer rating AI model.
Your input is a piece of reference text and a question.
You output is a score of whether the question is naturally written in {language_name} and whether it is answerable solely based on the reference text.

Output your score on a scale of 0-10.
A score of 0 should be given if the question is completely unanswerable based on the reference text or if the question is not written in fluent, natural {language_name}.
A score of 10 should be given if the question is fully answerable solely based on the refence text and the question is written in fluent, natural {language_name}.

Your output should be formatted like so:

### Question rating score
[YOUR SCORE]
\end{lstlisting}

\begin{lstlisting}[caption={Answer rating system message},captionpos=b,label={lst:aratingsysmsg}]
You are an answer rating AI model.
Your input is a piece of reference text, a question, and an answer.
You output is a score of how correct the answer is given the question and text.

Output your score on a scale of 0-10.
A score of 0 should be given if the answer is completely wrong based on the reference text or if the answer is not written in fluent, natural {language_name}.
A score of 10 should be given if the answer is completely correct based on the text and the answer is written in fluent, natural {language_name}.

Your output should be formatted like so:

### Answer rating score
[YOUR SCORE]
\end{lstlisting}

\begin{lstlisting}[caption={RAG system message},captionpos=b,label={lst:ragsysmsg}]
You are an retrival augmented generation (RAG) AI model.
Your input is a set of numbered documents and a question.
You output the id of the document(s) that best answer the question and then answer the question itself.

Your answer must be in fluent, natural {language_name}.

Your output should be formatted like so:

### Reference
[COMMA SEPARATED LIST OF RELEVANT DOCUMENT IDS]

### Answer
[YOUR ANSWER]
\end{lstlisting}

\begin{lstlisting}[caption={Answer checking system message for GPT4o},captionpos=b,label={lst:anschecksysmsg}]
You are a answer checking AI.
Given a context passage, a question, a correct reference answer, and a generated answer as inputs, determine whether the generated answer is correct based on the context given.

If the answer is not correct, output only FALSE.
If the answer is correct, output only TRUE.
\end{lstlisting}

\section{Training parameters}
\label{sec:trainingparams}

\begin{lstlisting}[caption={Selected training parameters for Axolotl LoRA training},captionpos=b,label={lst:trainingparams}]
sequence_len: 20000
sample_packing: false
eval_sample_packing: false
pad_to_sequence_len: true
adapter: lora
lora_r: 64
lora_alpha: 32
lora_dropout: 0.05
lora_target_linear: true
gradient_accumulation_steps: 1
micro_batch_size: 1
num_epochs: 1
optimizer: adamw_torch
lr_scheduler: cosine
learning_rate: 0.0002
train_on_inputs: false
group_by_length: false
bf16: auto
gradient_checkpointing: true
flash_attention: true
warmup_steps: 0
evals_per_epoch: 10
weight_decay: 0.0
\end{lstlisting}

\section{Extended results}
\label{sec:extendedresults}

\input{table_appendix_full_results}
\input{table_appendix_full_results2}

\end{document}

%% file: table_dataset.tex
\begin{table*}[t]
\centering
\begin{tabular}{|l|l|r|r|l|l|}
\hline
\textbf{Name} & \textbf{Domain} & \textbf{\# Texts} & \textbf{\# Questions} & \textbf{Language} & \textbf{Year} \\ \hline
\textbf{ARCD} & Wikipedia & 234 & 699 & Arabic & \citeyearpar{mozannar-etal-2019-neural} \\ \hline
\textbf{CalmQA} & QA site questions & 766 & 762 & 12 languages & \citeyearpar{arora2024calmqa} \\ \hline
\textbf{chaii-1} & Wikipedia & 110 & 125 & Hindi & \citeyearpar{chaii-hindi-and-tamil-question-answering} \\ \hline
\textbf{DRCD} & Wikipedia & 1,000 & 3,492 & Chinese & \citeyearpar{shao2018drcd} \\ \hline
\textbf{GermanQuAD} & Wikipedia & 474 & 2,203 & German & \citeyearpar{möller2021germanquad} \\ \hline
\textbf{JSQuAD} & Wikipedia & 1,145 & 4,429 & Japanese & \citeyearpar{kurihara2022jglue} \\ \hline
\textbf{KenSwQuAD} & Kencorpus & 1,157 & 5,978 & Swahili & \citeyearpar{wanjawa2023kenswquad} \\ \hline
\textbf{KorQuAD} & Wikipedia & 960 & 5,764 & Korean & \citeyearpar{lim2019korquad1} \\ \hline
\textbf{M2QA} & \begin{tabular}[c]{@{}l@{}}Reviews, news,\\ and creative writing\end{tabular} & 2,699 & 8,003 & 3 languages & \citeyearpar{englaender-etal-2024-m2qa} \\ \hline
\textbf{MLQA} & Wikipedia & 36,799 & 42,225 & 7 languages & \citeyearpar{lewis2019mlqa} \\ \hline
\textbf{NarrativeQA} & \begin{tabular}[c]{@{}l@{}}Book and movie\\ summaries\end{tabular} & 355 & 10,438 & English & \citeyearpar{kovcisky2018narrativeqa} \\ \hline
\textbf{PersianQA} & Wikipedia & 93 & 651 & Farsi & \citeyearpar{PersianQA} \\ \hline
\textbf{Pirá} & \begin{tabular}[c]{@{}l@{}}Environmental\\ reports\end{tabular} & 362 & 454 & 2 languages & \citeyearpar{pirozelli2024benchmarks} \\ \hline
\textbf{PublicHealth QA} & COVID FAQs & 886 & 886 & 8 languages & \citeyearpar{covidqa} \\ \hline
\textbf{SberQuAD} & Wikipedia & 3,971 & 5,036 & Russian & \citeyearpar{sberquad} \\ \hline
\textbf{SK-QuAD} & Wikipedia & 1,977 & 7,791 & Slovak & \citeyearpar{hladek2023slovak} \\ \hline
\textbf{SQAC} & Wikipedia & 634 & 1,908 & Spanish & \citeyearpar{maria} \\ \hline
\textbf{TQuad} & \begin{tabular}[c]{@{}l@{}}Islamic science\\ history articles\end{tabular} & 255 & 888 & Turkish & \citeyearpar{Peker2020} \\ \hline
\textbf{TyDi} & Wikipedia & 4,489 & 5,077 & 9 languages & \citeyearpar{clark2020tydi} \\ \hline
\textbf{XQuAD} & Wikipedia & 2,880 & 14,196 & 9 languages & \citeyearpar{Artetxe:etal:2019} \\ \hline
\end{tabular}
\caption{List of the data domains, number of unique texts and questions, number of languages, and the year of the publication for each dataset in our evaluation of ALoFTRAG.}
\label{tab:datasets}

\end{table*}

%% file: table_acc.tex
\begin{table*}[t]
\centering
\begin{tabular}{ll|rrrr|l|rrrr|}
\cline{3-6} \cline{8-11}
 &  & \multicolumn{4}{c|}{\textbf{Answer accuracy (\%)}} &  & \multicolumn{4}{c|}{\textbf{Reference accuracy (\%)}} \\ \cline{1-1} \cline{3-6} \cline{8-11} 
\multicolumn{1}{|l|}{} &  & \multicolumn{1}{l|}{\textbf{Base}} & \multicolumn{1}{l|}{\textbf{\begin{tabular}[c]{@{}l@{}}All\\ Steps\end{tabular}}} & \multicolumn{1}{l|}{\textbf{\begin{tabular}[c]{@{}l@{}}w/o\\ Step 1\end{tabular}}} & \multicolumn{1}{l|}{\textbf{\begin{tabular}[c]{@{}l@{}}w/o\\ Step 3\end{tabular}}} &  & \multicolumn{1}{l|}{\textbf{Base}} & \multicolumn{1}{l|}{\textbf{\begin{tabular}[c]{@{}l@{}}All\\ Steps\end{tabular}}} & \multicolumn{1}{l|}{\textbf{\begin{tabular}[c]{@{}l@{}}w/o\\ Step 1\end{tabular}}} & \multicolumn{1}{l|}{\textbf{\begin{tabular}[c]{@{}l@{}}w/o\\ Step 3\end{tabular}}} \\ \cline{1-1} \cline{3-6} \cline{8-11} 
\multicolumn{1}{|l|}{ARCD} &  & \multicolumn{1}{r|}{78.0} & \multicolumn{1}{r|}{78.8} & \multicolumn{1}{r|}{77.8} & \textbf{79.3} &  & \multicolumn{1}{r|}{72.0} & \multicolumn{1}{r|}{72.2} & \multicolumn{1}{r|}{73.1} & \textbf{77.4} \\ \cline{1-1} \cline{3-6} \cline{8-11} 
\multicolumn{1}{|l|}{CalmQA} &  & \multicolumn{1}{r|}{60.8} & \multicolumn{1}{r|}{\textbf{68.1}} & \multicolumn{1}{r|}{66.0} & 65.7 &  & \multicolumn{1}{r|}{71.6} & \multicolumn{1}{r|}{86.1} & \multicolumn{1}{r|}{84.0} & \textbf{89.0} \\ \cline{1-1} \cline{3-6} \cline{8-11} 
\multicolumn{1}{|l|}{chaii-1} &  & \multicolumn{1}{r|}{92.0} & \multicolumn{1}{r|}{89.6} & \multicolumn{1}{r|}{92.0} & \textbf{95.2} &  & \multicolumn{1}{r|}{83.2} & \multicolumn{1}{r|}{89.6} & \multicolumn{1}{r|}{89.6} & \textbf{93.6} \\ \cline{1-1} \cline{3-6} \cline{8-11} 
\multicolumn{1}{|l|}{DRCD} &  & \multicolumn{1}{r|}{88.1} & \multicolumn{1}{r|}{89.5} & \multicolumn{1}{r|}{88.7} & \textbf{90.4} &  & \multicolumn{1}{r|}{63.7} & \multicolumn{1}{r|}{82.1} & \multicolumn{1}{r|}{83.7} & \textbf{89.3} \\ \cline{1-1} \cline{3-6} \cline{8-11} 
\multicolumn{1}{|l|}{GermanQuAD} &  & \multicolumn{1}{r|}{82.9} & \multicolumn{1}{r|}{83.8} & \multicolumn{1}{r|}{82.1} & \textbf{84.3} &  & \multicolumn{1}{r|}{52.7} & \multicolumn{1}{r|}{78.9} & \multicolumn{1}{r|}{77.2} & \textbf{85.2} \\ \cline{1-1} \cline{3-6} \cline{8-11} 
\multicolumn{1}{|l|}{JSQuAD} &  & \multicolumn{1}{r|}{84.2} & \multicolumn{1}{r|}{\textbf{88.7}} & \multicolumn{1}{r|}{87.6} & 88.6 &  & \multicolumn{1}{r|}{81.3} & \multicolumn{1}{r|}{89.5} & \multicolumn{1}{r|}{89.3} & \textbf{91.3} \\ \cline{1-1} \cline{3-6} \cline{8-11} 
\multicolumn{1}{|l|}{KenSwQuAD} &  & \multicolumn{1}{r|}{26.1} & \multicolumn{1}{r|}{33.1} & \multicolumn{1}{r|}{32.5} & \textbf{33.7} &  & \multicolumn{1}{r|}{\textbf{54.6}} & \multicolumn{1}{r|}{32.1} & \multicolumn{1}{r|}{35.6} & 41.4 \\ \cline{1-1} \cline{3-6} \cline{8-11} 
\multicolumn{1}{|l|}{KorQuAD} &  & \multicolumn{1}{r|}{84.8} & \multicolumn{1}{r|}{87.7} & \multicolumn{1}{r|}{85.8} & \textbf{88.6} &  & \multicolumn{1}{r|}{61.0} & \multicolumn{1}{r|}{84.1} & \multicolumn{1}{r|}{83.1} & \textbf{89.6} \\ \cline{1-1} \cline{3-6} \cline{8-11} 
\multicolumn{1}{|l|}{M2QA} &  & \multicolumn{1}{r|}{50.3} & \multicolumn{1}{r|}{\textbf{58.0}} & \multicolumn{1}{r|}{55.6} & 57.6 &  & \multicolumn{1}{r|}{56.6} & \multicolumn{1}{r|}{58.4} & \multicolumn{1}{r|}{60.5} & \textbf{63.0} \\ \cline{1-1} \cline{3-6} \cline{8-11} 
\multicolumn{1}{|l|}{MLQA} &  & \multicolumn{1}{r|}{58.0} & \multicolumn{1}{r|}{61.1} & \multicolumn{1}{r|}{59.5} & \textbf{61.8} &  & \multicolumn{1}{r|}{58.8} & \multicolumn{1}{r|}{65.2} & \multicolumn{1}{r|}{66.7} & \textbf{69.0} \\ \cline{1-1} \cline{3-6} \cline{8-11} 
\multicolumn{1}{|l|}{NarrativeQA} &  & \multicolumn{1}{r|}{79.1} & \multicolumn{1}{r|}{80.9} & \multicolumn{1}{r|}{80.2} & \textbf{81.8} &  & \multicolumn{1}{r|}{54.7} & \multicolumn{1}{r|}{70.5} & \multicolumn{1}{r|}{70.8} & \textbf{81.3} \\ \cline{1-1} \cline{3-6} \cline{8-11} 
\multicolumn{1}{|l|}{PersianQA} &  & \multicolumn{1}{r|}{80.5} & \multicolumn{1}{r|}{82.9} & \multicolumn{1}{r|}{82.9} & \textbf{87.4} &  & \multicolumn{1}{r|}{80.6} & \multicolumn{1}{r|}{92.5} & \multicolumn{1}{r|}{94.5} & \textbf{96.8} \\ \cline{1-1} \cline{3-6} \cline{8-11} 
\multicolumn{1}{|l|}{Pirá} &  & \multicolumn{1}{r|}{74.0} & \multicolumn{1}{r|}{\textbf{78.6}} & \multicolumn{1}{r|}{76.0} & 77.5 &  & \multicolumn{1}{r|}{75.5} & \multicolumn{1}{r|}{83.0} & \multicolumn{1}{r|}{85.0} & \textbf{86.6} \\ \cline{1-1} \cline{3-6} \cline{8-11} 
\multicolumn{1}{|l|}{PublicHealth QA} &  & \multicolumn{1}{r|}{68.2} & \multicolumn{1}{r|}{73.6} & \multicolumn{1}{r|}{\textbf{75.6}} & 75.3 &  & \multicolumn{1}{r|}{82.7} & \multicolumn{1}{r|}{68.8} & \multicolumn{1}{r|}{73.8} & \textbf{74.6} \\ \cline{1-1} \cline{3-6} \cline{8-11} 
\multicolumn{1}{|l|}{SberQuAD} &  & \multicolumn{1}{r|}{86.6} & \multicolumn{1}{r|}{88.6} & \multicolumn{1}{r|}{87.3} & \textbf{89.0} &  & \multicolumn{1}{r|}{80.0} & \multicolumn{1}{r|}{88.5} & \multicolumn{1}{r|}{89.9} & \textbf{92.1} \\ \cline{1-1} \cline{3-6} \cline{8-11} 
\multicolumn{1}{|l|}{SK-QuAD} &  & \multicolumn{1}{r|}{84.6} & \multicolumn{1}{r|}{86.9} & \multicolumn{1}{r|}{85.2} & \textbf{87.6} &  & \multicolumn{1}{r|}{71.3} & \multicolumn{1}{r|}{86.5} & \multicolumn{1}{r|}{88.0} & \textbf{92.1} \\ \cline{1-1} \cline{3-6} \cline{8-11} 
\multicolumn{1}{|l|}{SQAC} &  & \multicolumn{1}{r|}{77.7} & \multicolumn{1}{r|}{\textbf{80.5}} & \multicolumn{1}{r|}{77.6} & 79.7 &  & \multicolumn{1}{r|}{67.0} & \multicolumn{1}{r|}{77.1} & \multicolumn{1}{r|}{78.7} & \textbf{81.4} \\ \cline{1-1} \cline{3-6} \cline{8-11} 
\multicolumn{1}{|l|}{TQuad} &  & \multicolumn{1}{r|}{79.6} & \multicolumn{1}{r|}{81.6} & \multicolumn{1}{r|}{82.5} & \textbf{82.9} &  & \multicolumn{1}{r|}{63.7} & \multicolumn{1}{r|}{72.1} & \multicolumn{1}{r|}{75.8} & \textbf{79.5} \\ \cline{1-1} \cline{3-6} \cline{8-11} 
\multicolumn{1}{|l|}{TyDi} &  & \multicolumn{1}{r|}{83.5} & \multicolumn{1}{r|}{84.3} & \multicolumn{1}{r|}{83.9} & \textbf{84.9} &  & \multicolumn{1}{r|}{83.5} & \multicolumn{1}{r|}{89.6} & \multicolumn{1}{r|}{91.5} & \textbf{92.1} \\ \cline{1-1} \cline{3-6} \cline{8-11} 
\multicolumn{1}{|l|}{XQuAD} &  & \multicolumn{1}{r|}{81.1} & \multicolumn{1}{r|}{\textbf{83.2}} & \multicolumn{1}{r|}{81.4} & 82.9 &  & \multicolumn{1}{r|}{73.2} & \multicolumn{1}{r|}{87.9} & \multicolumn{1}{r|}{89.7} & \textbf{92.1} \\ \cline{1-1} \cline{3-6} \cline{8-11} 
\multicolumn{1}{|l|}{\textbf{Mean}} &  & \multicolumn{1}{r|}{75.0} & \multicolumn{1}{r|}{78.0} & \multicolumn{1}{r|}{77.0} & 78.7 &  & \multicolumn{1}{r|}{69.4} & \multicolumn{1}{r|}{77.7} & \multicolumn{1}{r|}{79.0} & 82.9 \\ \cline{1-1} \cline{3-6} \cline{8-11} 
\end{tabular}
\caption{Percentage answer and citation accuracy averaged over all language subsets within each dataset for the base model and the ALoFTRAG models and ablation tests.}
\label{tab:acc_results_dataset_averaged}
\end{table*}

%% file: table_hard_easy_acc.tex
\begin{table}
\centering
\begin{tabular}{llrrrr}
\cline{3-6}
\textbf{} & \multicolumn{1}{l|}{} & \multicolumn{1}{l|}{\textbf{Base}} & \multicolumn{1}{l|}{\textbf{\begin{tabular}[c]{@{}l@{}}All\\ Steps\end{tabular}}} & \multicolumn{1}{l|}{\textbf{\begin{tabular}[c]{@{}l@{}}w/o\\ Step 1\end{tabular}}} & \multicolumn{1}{l|}{\textbf{\begin{tabular}[c]{@{}l@{}}w/o\\ Step 3\end{tabular}}} \\ \cline{3-6} 
 &  & \multicolumn{1}{l}{} & \multicolumn{1}{l}{} & \multicolumn{1}{l}{} & \multicolumn{1}{l}{} \\ \hline

\multicolumn{1}{|l|}{\multirow{2}{*}{\textbf{\begin{tabular}[c]{@{}l@{}}Ans.\\ Acc.\end{tabular}}}} & \multicolumn{1}{l|}{\textbf{Easy}} & \multicolumn{1}{r|}{76.9} & \multicolumn{1}{r|}{79.8} & \multicolumn{1}{r|}{78.8} & \multicolumn{1}{r|}{\textbf{80.6}} \\ \cline{2-6} 
\multicolumn{1}{|l|}{} & \multicolumn{1}{l|}{\textbf{Hard}} & \multicolumn{1}{r|}{48.2} & \multicolumn{1}{r|}{54.5} & \multicolumn{1}{r|}{53.8} & \multicolumn{1}{r|}{\textbf{55.2}} \\ \hline
 &  & \multicolumn{1}{l}{} & \multicolumn{1}{l}{} & \multicolumn{1}{l}{} & \multicolumn{1}{l}{} \\ \hline

\multicolumn{1}{|l|}{\multirow{2}{*}{\textbf{\begin{tabular}[c]{@{}l@{}}Ref.\\ Acc.\end{tabular}}}} & \multicolumn{1}{l|}{\textbf{Easy}} & \multicolumn{1}{r|}{70.7} & \multicolumn{1}{r|}{80.0} & \multicolumn{1}{r|}{81.3} & \multicolumn{1}{r|}{\textbf{85.3}} \\ \cline{2-6} 
\multicolumn{1}{|l|}{} & \multicolumn{1}{l|}{\textbf{Hard}} & \multicolumn{1}{r|}{41.0} & \multicolumn{1}{r|}{44.5} & \multicolumn{1}{r|}{48.4} & \multicolumn{1}{r|}{\textbf{49.7}} \\ \hline
\end{tabular}
\caption{Percentage answer and reference accuracy for both easy and hard questions averaged across all datasets.}
\label{tab:hard_easy_results}
\end{table}

%% file: table_appendix_full_results.tex
\begin{table*}
\centering
\begin{tabular}{lll|rrrr|l|rrrr|}
\cline{4-7} \cline{9-12}
 &  &  & \multicolumn{4}{l|}{\textbf{Answer accuracy (\%)}} &  & \multicolumn{4}{l|}{\textbf{Reference Accuracy (\%)}} \\ \cline{1-2} \cline{4-7} \cline{9-12} 
\multicolumn{1}{|l|}{\textbf{Dataset name}} & \multicolumn{1}{l|}{\textbf{Language}} &  & \multicolumn{1}{l|}{\textbf{Base}} & \multicolumn{1}{l|}{\textbf{\begin{tabular}[c]{@{}l@{}}All\\ Steps\end{tabular}}} & \multicolumn{1}{l|}{\textbf{\begin{tabular}[c]{@{}l@{}}w/o\\ Step 1\end{tabular}}} & \multicolumn{1}{l|}{\textbf{\begin{tabular}[c]{@{}l@{}}w/o\\ Step 3\end{tabular}}} &  & \multicolumn{1}{l|}{\textbf{Base}} & \multicolumn{1}{l|}{\textbf{\begin{tabular}[c]{@{}l@{}}All\\ Steps\end{tabular}}} & \multicolumn{1}{l|}{\textbf{\begin{tabular}[c]{@{}l@{}}w/o\\ Step 1\end{tabular}}} & \multicolumn{1}{l|}{\textbf{\begin{tabular}[c]{@{}l@{}}w/o\\ Step 3\end{tabular}}} \\ \cline{1-2} \cline{4-7} \cline{9-12} 
\multicolumn{1}{|l|}{ARCD} & \multicolumn{1}{l|}{Arabic} &  & \multicolumn{1}{r|}{78.0} & \multicolumn{1}{r|}{78.8} & \multicolumn{1}{r|}{77.8} & \textbf{79.3} &  & \multicolumn{1}{r|}{72.0} & \multicolumn{1}{r|}{72.2} & \multicolumn{1}{r|}{73.1} & \textbf{77.4} \\ \cline{1-2} \cline{4-7} \cline{9-12} 
\multicolumn{1}{|l|}{\multirow{12}{*}{CalmQA}} & \multicolumn{1}{l|}{Arabic} &  & \multicolumn{1}{r|}{80.3} & \multicolumn{1}{r|}{\textbf{83.6}} & \multicolumn{1}{r|}{74.2} & 82.3 &  & \multicolumn{1}{r|}{80.3} & \multicolumn{1}{r|}{95.1} & \multicolumn{1}{r|}{93.5} & \textbf{95.2} \\ \cline{2-2} \cline{4-7} \cline{9-12} 
\multicolumn{1}{|l|}{} & \multicolumn{1}{l|}{Chinese} &  & \multicolumn{1}{r|}{33.9} & \multicolumn{1}{r|}{\textbf{53.6}} & \multicolumn{1}{r|}{46.4} & 46.4 &  & \multicolumn{1}{r|}{66.1} & \multicolumn{1}{r|}{67.9} & \multicolumn{1}{r|}{66.1} & \textbf{73.2} \\ \cline{2-2} \cline{4-7} \cline{9-12} 
\multicolumn{1}{|l|}{} & \multicolumn{1}{l|}{English} &  & \multicolumn{1}{r|}{79.8} & \multicolumn{1}{r|}{85.1} & \multicolumn{1}{r|}{\textbf{88.4}} & 86.5 &  & \multicolumn{1}{r|}{86.2} & \multicolumn{1}{r|}{91.5} & \multicolumn{1}{r|}{92.6} & \textbf{94.8} \\ \cline{2-2} \cline{4-7} \cline{9-12} 
\multicolumn{1}{|l|}{} & \multicolumn{1}{l|}{German} &  & \multicolumn{1}{r|}{68.1} & \multicolumn{1}{r|}{68.1} & \multicolumn{1}{r|}{73.2} & \textbf{73.9} &  & \multicolumn{1}{r|}{83.3} & \multicolumn{1}{r|}{88.9} & \multicolumn{1}{r|}{88.7} & \textbf{94.2} \\ \cline{2-2} \cline{4-7} \cline{9-12} 
\multicolumn{1}{|l|}{} & \multicolumn{1}{l|}{Hebrew} &  & \multicolumn{1}{r|}{53.7} & \multicolumn{1}{r|}{\textbf{56.7}} & \multicolumn{1}{r|}{55.1} & 51.4 &  & \multicolumn{1}{r|}{56.7} & \multicolumn{1}{r|}{83.6} & \multicolumn{1}{r|}{79.7} & \textbf{85.7} \\ \cline{2-2} \cline{4-7} \cline{9-12} 
\multicolumn{1}{|l|}{} & \multicolumn{1}{l|}{Hindi} &  & \multicolumn{1}{r|}{\textbf{86.7}} & \multicolumn{1}{r|}{80.0} & \multicolumn{1}{r|}{84.1} & 82.8 &  & \multicolumn{1}{r|}{56.7} & \multicolumn{1}{r|}{86.7} & \multicolumn{1}{r|}{\textbf{92.1}} & 90.6 \\ \cline{2-2} \cline{4-7} \cline{9-12} 
\multicolumn{1}{|l|}{} & \multicolumn{1}{l|}{Hungarian} &  & \multicolumn{1}{r|}{26.8} & \multicolumn{1}{r|}{53.6} & \multicolumn{1}{r|}{44.6} & \textbf{57.1} &  & \multicolumn{1}{r|}{69.6} & \multicolumn{1}{r|}{76.8} & \multicolumn{1}{r|}{64.3} & \textbf{80.4} \\ \cline{2-2} \cline{4-7} \cline{9-12} 
\multicolumn{1}{|l|}{} & \multicolumn{1}{l|}{Japanese} &  & \multicolumn{1}{r|}{62.5} & \multicolumn{1}{r|}{\textbf{71.4}} & \multicolumn{1}{r|}{64.3} & 58.9 &  & \multicolumn{1}{r|}{80.4} & \multicolumn{1}{r|}{91.1} & \multicolumn{1}{r|}{83.9} & \textbf{94.6} \\ \cline{2-2} \cline{4-7} \cline{9-12} 
\multicolumn{1}{|l|}{} & \multicolumn{1}{l|}{Kirundi} &  & \multicolumn{1}{r|}{13.2} & \multicolumn{1}{r|}{\textbf{18.4}} & \multicolumn{1}{r|}{\textbf{18.4}} & 7.7 &  & \multicolumn{1}{r|}{55.3} & \multicolumn{1}{r|}{76.3} & \multicolumn{1}{r|}{78.9} & \textbf{82.1} \\ \cline{2-2} \cline{4-7} \cline{9-12} 
\multicolumn{1}{|l|}{} & \multicolumn{1}{l|}{Korean} &  & \multicolumn{1}{r|}{83.9} & \multicolumn{1}{r|}{83.9} & \multicolumn{1}{r|}{\textbf{89.1}} & 87.5 &  & \multicolumn{1}{r|}{82.1} & \multicolumn{1}{r|}{\textbf{98.2}} & \multicolumn{1}{r|}{96.4} & \textbf{98.2} \\ \cline{2-2} \cline{4-7} \cline{9-12} 
\multicolumn{1}{|l|}{} & \multicolumn{1}{l|}{Russian} &  & \multicolumn{1}{r|}{72.5} & \multicolumn{1}{r|}{82.4} & \multicolumn{1}{r|}{\textbf{83.0}} & 77.8 &  & \multicolumn{1}{r|}{60.8} & \multicolumn{1}{r|}{92.2} & \multicolumn{1}{r|}{90.6} & \textbf{94.4} \\ \cline{2-2} \cline{4-7} \cline{9-12} 
\multicolumn{1}{|l|}{} & \multicolumn{1}{l|}{Spanish} &  & \multicolumn{1}{r|}{68.5} & \multicolumn{1}{r|}{\textbf{80.8}} & \multicolumn{1}{r|}{70.8} & 75.7 &  & \multicolumn{1}{r|}{82.2} & \multicolumn{1}{r|}{84.9} & \multicolumn{1}{r|}{80.6} & \textbf{85.1} \\ \cline{1-2} \cline{4-7} \cline{9-12} 
\multicolumn{1}{|l|}{chaii-1} & \multicolumn{1}{l|}{Hindi} &  & \multicolumn{1}{r|}{92.0} & \multicolumn{1}{r|}{89.6} & \multicolumn{1}{r|}{92.0} & \textbf{95.2} &  & \multicolumn{1}{r|}{83.2} & \multicolumn{1}{r|}{89.6} & \multicolumn{1}{r|}{89.6} & \textbf{93.6} \\ \cline{1-2} \cline{4-7} \cline{9-12} 
\multicolumn{1}{|l|}{DRCD} & \multicolumn{1}{l|}{Chinese} &  & \multicolumn{1}{r|}{88.1} & \multicolumn{1}{r|}{89.5} & \multicolumn{1}{r|}{88.7} & \textbf{90.4} &  & \multicolumn{1}{r|}{63.7} & \multicolumn{1}{r|}{82.1} & \multicolumn{1}{r|}{83.7} & \textbf{89.3} \\ \cline{1-2} \cline{4-7} \cline{9-12} 
\multicolumn{1}{|l|}{GermanQuAD} & \multicolumn{1}{l|}{German} &  & \multicolumn{1}{r|}{82.9} & \multicolumn{1}{r|}{83.8} & \multicolumn{1}{r|}{82.1} & \textbf{84.3} &  & \multicolumn{1}{r|}{52.7} & \multicolumn{1}{r|}{78.9} & \multicolumn{1}{r|}{77.2} & \textbf{85.2} \\ \cline{1-2} \cline{4-7} \cline{9-12} 
\multicolumn{1}{|l|}{JSQuAD} & \multicolumn{1}{l|}{Japanese} &  & \multicolumn{1}{r|}{84.2} & \multicolumn{1}{r|}{\textbf{88.7}} & \multicolumn{1}{r|}{87.6} & 88.6 &  & \multicolumn{1}{r|}{81.3} & \multicolumn{1}{r|}{89.5} & \multicolumn{1}{r|}{89.3} & \textbf{91.3} \\ \cline{1-2} \cline{4-7} \cline{9-12} 
\multicolumn{1}{|l|}{KenSwQuAD} & \multicolumn{1}{l|}{Swahili} &  & \multicolumn{1}{r|}{26.1} & \multicolumn{1}{r|}{33.1} & \multicolumn{1}{r|}{32.5} & \textbf{33.7} &  & \multicolumn{1}{r|}{\textbf{54.6}} & \multicolumn{1}{r|}{32.1} & \multicolumn{1}{r|}{35.6} & 41.4 \\ \cline{1-2} \cline{4-7} \cline{9-12} 
\multicolumn{1}{|l|}{KorQuAD} & \multicolumn{1}{l|}{Korean} &  & \multicolumn{1}{r|}{84.8} & \multicolumn{1}{r|}{87.7} & \multicolumn{1}{r|}{85.8} & \textbf{88.6} &  & \multicolumn{1}{r|}{61.0} & \multicolumn{1}{r|}{84.1} & \multicolumn{1}{r|}{83.1} & \textbf{89.6} \\ \cline{1-2} \cline{4-7} \cline{9-12} 
\multicolumn{1}{|l|}{\multirow{3}{*}{M2QA}} & \multicolumn{1}{l|}{Chinese} &  & \multicolumn{1}{r|}{53.1} & \multicolumn{1}{r|}{\textbf{65.0}} & \multicolumn{1}{r|}{61.3} & 63.5 &  & \multicolumn{1}{r|}{\textbf{63.8}} & \multicolumn{1}{r|}{60.1} & \multicolumn{1}{r|}{61.4} & 63.4 \\ \cline{2-2} \cline{4-7} \cline{9-12} 
\multicolumn{1}{|l|}{} & \multicolumn{1}{l|}{German} &  & \multicolumn{1}{r|}{54.1} & \multicolumn{1}{r|}{\textbf{60.3}} & \multicolumn{1}{r|}{58.0} & 60.2 &  & \multicolumn{1}{r|}{56.6} & \multicolumn{1}{r|}{58.5} & \multicolumn{1}{r|}{61.6} & \textbf{63.6} \\ \cline{2-2} \cline{4-7} \cline{9-12} 
\multicolumn{1}{|l|}{} & \multicolumn{1}{l|}{Turkish} &  & \multicolumn{1}{r|}{43.8} & \multicolumn{1}{r|}{48.8} & \multicolumn{1}{r|}{47.4} & \textbf{49.2} &  & \multicolumn{1}{r|}{49.5} & \multicolumn{1}{r|}{56.7} & \multicolumn{1}{r|}{58.5} & \textbf{61.9} \\ \cline{1-2} \cline{4-7} \cline{9-12} 
\multicolumn{1}{|l|}{\multirow{7}{*}{MLQA}} & \multicolumn{1}{l|}{Arabic} &  & \multicolumn{1}{r|}{50.5} & \multicolumn{1}{r|}{54.3} & \multicolumn{1}{r|}{51.4} & \textbf{55.8} &  & \multicolumn{1}{r|}{55.1} & \multicolumn{1}{r|}{62.2} & \multicolumn{1}{r|}{62.5} & \textbf{66.1} \\ \cline{2-2} \cline{4-7} \cline{9-12} 
\multicolumn{1}{|l|}{} & \multicolumn{1}{l|}{Chinese} &  & \multicolumn{1}{r|}{59.8} & \multicolumn{1}{r|}{63.7} & \multicolumn{1}{r|}{61.9} & \textbf{63.8} &  & \multicolumn{1}{r|}{62.8} & \multicolumn{1}{r|}{68.5} & \multicolumn{1}{r|}{69.6} & \textbf{71.4} \\ \cline{2-2} \cline{4-7} \cline{9-12} 
\multicolumn{1}{|l|}{} & \multicolumn{1}{l|}{English} &  & \multicolumn{1}{r|}{68.2} & \multicolumn{1}{r|}{72.0} & \multicolumn{1}{r|}{70.9} & \textbf{72.3} &  & \multicolumn{1}{r|}{64.3} & \multicolumn{1}{r|}{70.5} & \multicolumn{1}{r|}{71.8} & \textbf{74.2} \\ \cline{2-2} \cline{4-7} \cline{9-12} 
\multicolumn{1}{|l|}{} & \multicolumn{1}{l|}{German} &  & \multicolumn{1}{r|}{56.4} & \multicolumn{1}{r|}{59.0} & \multicolumn{1}{r|}{58.2} & \textbf{60.1} &  & \multicolumn{1}{r|}{58.7} & \multicolumn{1}{r|}{63.6} & \multicolumn{1}{r|}{66.6} & \textbf{67.2} \\ \cline{2-2} \cline{4-7} \cline{9-12} 
\multicolumn{1}{|l|}{} & \multicolumn{1}{l|}{Hindi} &  & \multicolumn{1}{r|}{48.3} & \multicolumn{1}{r|}{51.3} & \multicolumn{1}{r|}{49.7} & \textbf{52.1} &  & \multicolumn{1}{r|}{50.7} & \multicolumn{1}{r|}{57.6} & \multicolumn{1}{r|}{58.5} & \textbf{62.9} \\ \cline{2-2} \cline{4-7} \cline{9-12} 
\multicolumn{1}{|l|}{} & \multicolumn{1}{l|}{Spanish} &  & \multicolumn{1}{r|}{63.4} & \multicolumn{1}{r|}{\textbf{66.2}} & \multicolumn{1}{r|}{64.2} & \textbf{66.2} &  & \multicolumn{1}{r|}{65.2} & \multicolumn{1}{r|}{70.2} & \multicolumn{1}{r|}{71.4} & \textbf{72.8} \\ \cline{2-2} \cline{4-7} \cline{9-12} 
\multicolumn{1}{|l|}{} & \multicolumn{1}{l|}{Vietnamese} &  & \multicolumn{1}{r|}{59.5} & \multicolumn{1}{r|}{61.5} & \multicolumn{1}{r|}{60.1} & \textbf{62.4} &  & \multicolumn{1}{r|}{54.8} & \multicolumn{1}{r|}{64.0} & \multicolumn{1}{r|}{66.4} & \textbf{68.3} \\ \cline{1-2} \cline{4-7} \cline{9-12} 
\multicolumn{1}{|l|}{NarrativeQA} & \multicolumn{1}{l|}{English} &  & \multicolumn{1}{r|}{79.1} & \multicolumn{1}{r|}{80.9} & \multicolumn{1}{r|}{80.2} & \textbf{81.8} &  & \multicolumn{1}{r|}{54.7} & \multicolumn{1}{r|}{70.5} & \multicolumn{1}{r|}{70.8} & \textbf{81.3} \\ \cline{1-2} \cline{4-7} \cline{9-12} 
\multicolumn{1}{|l|}{PersianQA} & \multicolumn{1}{l|}{Persian} &  & \multicolumn{1}{r|}{80.5} & \multicolumn{1}{r|}{82.9} & \multicolumn{1}{r|}{82.9} & \textbf{87.4} &  & \multicolumn{1}{r|}{80.6} & \multicolumn{1}{r|}{92.5} & \multicolumn{1}{r|}{94.5} & \textbf{96.8} \\ \cline{1-2} \cline{4-7} \cline{9-12} 
\multicolumn{1}{|l|}{\multirow{2}{*}{Pirá}} & \multicolumn{1}{l|}{English} &  & \multicolumn{1}{r|}{77.9} & \multicolumn{1}{r|}{80.5} & \multicolumn{1}{r|}{78.0} & \textbf{81.9} &  & \multicolumn{1}{r|}{81.0} & \multicolumn{1}{r|}{83.6} & \multicolumn{1}{r|}{87.2} & \textbf{87.7} \\ \cline{2-2} \cline{4-7} \cline{9-12} 
\multicolumn{1}{|l|}{} & \multicolumn{1}{l|}{Portuguese} &  & \multicolumn{1}{r|}{70.0} & \multicolumn{1}{r|}{\textbf{76.7}} & \multicolumn{1}{r|}{74.0} & 73.1 &  & \multicolumn{1}{r|}{70.0} & \multicolumn{1}{r|}{82.4} & \multicolumn{1}{r|}{82.8} & \textbf{85.5} \\ \cline{1-2} \cline{4-7} \cline{9-12} 
\multicolumn{1}{|l|}{\multirow{8}{*}{\begin{tabular}[c]{@{}l@{}}PublicHealth\\ QA\end{tabular}}} & \multicolumn{1}{l|}{Arabic} &  & \multicolumn{1}{r|}{68.6} & \multicolumn{1}{r|}{73.3} & \multicolumn{1}{r|}{75.6} & \textbf{76.7} &  & \multicolumn{1}{r|}{\textbf{82.6}} & \multicolumn{1}{r|}{73.3} & \multicolumn{1}{r|}{76.7} & 81.4 \\ \cline{2-2} \cline{4-7} \cline{9-12} 
\multicolumn{1}{|l|}{} & \multicolumn{1}{l|}{Chinese} &  & \multicolumn{1}{r|}{60.7} & \multicolumn{1}{r|}{\textbf{76.1}} & \multicolumn{1}{r|}{73.0} & 74.2 &  & \multicolumn{1}{r|}{\textbf{84.0}} & \multicolumn{1}{r|}{69.3} & \multicolumn{1}{r|}{69.3} & 69.9 \\ \cline{2-2} \cline{4-7} \cline{9-12} 
\multicolumn{1}{|l|}{} & \multicolumn{1}{l|}{English} &  & \multicolumn{1}{r|}{67.3} & \multicolumn{1}{r|}{75.6} & \multicolumn{1}{r|}{\textbf{76.6}} & 73.8 &  & \multicolumn{1}{r|}{\textbf{87.5}} & \multicolumn{1}{r|}{69.0} & \multicolumn{1}{r|}{73.7} & 71.4 \\ \cline{2-2} \cline{4-7} \cline{9-12} 
\multicolumn{1}{|l|}{} & \multicolumn{1}{l|}{French} &  & \multicolumn{1}{r|}{75.0} & \multicolumn{1}{r|}{78.8} & \multicolumn{1}{r|}{\textbf{80.8}} & 78.5 &  & \multicolumn{1}{r|}{77.5} & \multicolumn{1}{r|}{75.0} & \multicolumn{1}{r|}{\textbf{79.5}} & 78.5 \\ \cline{2-2} \cline{4-7} \cline{9-12} 
\multicolumn{1}{|l|}{} & \multicolumn{1}{l|}{Korean} &  & \multicolumn{1}{r|}{63.6} & \multicolumn{1}{r|}{68.8} & \multicolumn{1}{r|}{\textbf{71.4}} & \textbf{71.4} &  & \multicolumn{1}{r|}{\textbf{85.7}} & \multicolumn{1}{r|}{59.7} & \multicolumn{1}{r|}{66.2} & 68.8 \\ \cline{2-2} \cline{4-7} \cline{9-12} 
\multicolumn{1}{|l|}{} & \multicolumn{1}{l|}{Russian} &  & \multicolumn{1}{r|}{74.6} & \multicolumn{1}{r|}{74.6} & \multicolumn{1}{r|}{\textbf{80.0}} & \textbf{80.0} &  & \multicolumn{1}{r|}{\textbf{82.5}} & \multicolumn{1}{r|}{73.0} & \multicolumn{1}{r|}{78.5} & 80.0 \\ \cline{2-2} \cline{4-7} \cline{9-12} 
\multicolumn{1}{|l|}{} & \multicolumn{1}{l|}{Spanish} &  & \multicolumn{1}{r|}{65.6} & \multicolumn{1}{r|}{72.5} & \multicolumn{1}{r|}{\textbf{78.6}} & 75.2 &  & \multicolumn{1}{r|}{\textbf{83.8}} & \multicolumn{1}{r|}{66.2} & \multicolumn{1}{r|}{73.6} & 72.6 \\ \cline{2-2} \cline{4-7} \cline{9-12} 
\multicolumn{1}{|l|}{} & \multicolumn{1}{l|}{Vietnamese} &  & \multicolumn{1}{r|}{70.1} & \multicolumn{1}{r|}{68.8} & \multicolumn{1}{r|}{68.8} & \textbf{72.7} &  & \multicolumn{1}{r|}{\textbf{77.9}} & \multicolumn{1}{r|}{64.9} & \multicolumn{1}{r|}{72.7} & 74.0 \\ \cline{1-2} \cline{4-7} \cline{9-12} 
\multicolumn{1}{|l|}{SberQuAD} & \multicolumn{1}{l|}{Russian} &  & \multicolumn{1}{r|}{86.6} & \multicolumn{1}{r|}{88.6} & \multicolumn{1}{r|}{87.3} & \textbf{89.0} &  & \multicolumn{1}{r|}{80.0} & \multicolumn{1}{r|}{88.5} & \multicolumn{1}{r|}{89.9} & \textbf{92.1} \\ \cline{1-2} \cline{4-7} \cline{9-12} 
\multicolumn{1}{|l|}{SK-QuAD} & \multicolumn{1}{l|}{Slovak} &  & \multicolumn{1}{r|}{84.6} & \multicolumn{1}{r|}{86.9} & \multicolumn{1}{r|}{85.2} & \textbf{87.6} &  & \multicolumn{1}{r|}{71.3} & \multicolumn{1}{r|}{86.5} & \multicolumn{1}{r|}{88.0} & \textbf{92.1} \\ \cline{1-2} \cline{4-7} \cline{9-12} 
\multicolumn{1}{|l|}{SQAC} & \multicolumn{1}{l|}{Spanish} &  & \multicolumn{1}{r|}{77.7} & \multicolumn{1}{r|}{\textbf{80.5}} & \multicolumn{1}{r|}{77.6} & 79.7 &  & \multicolumn{1}{r|}{67.0} & \multicolumn{1}{r|}{77.1} & \multicolumn{1}{r|}{78.7} & \textbf{81.4} \\ \cline{1-2} \cline{4-7} \cline{9-12} 
\multicolumn{1}{|l|}{TQuad} & \multicolumn{1}{l|}{Turkish} &  & \multicolumn{1}{r|}{79.6} & \multicolumn{1}{r|}{81.6} & \multicolumn{1}{r|}{82.5} & \textbf{82.9} &  & \multicolumn{1}{r|}{63.7} & \multicolumn{1}{r|}{72.1} & \multicolumn{1}{r|}{75.8} & \textbf{79.5} \\ \cline{1-2} \cline{4-7} \cline{9-12} 
\end{tabular}
\caption{Full per language answer and reference accuracies for each dataset. (Continued on the next page)}
\label{tab:fullresultsappendix}
\end{table*}

%% file: table_appendix_full_results2.tex
\begin{table*}[t]
\centering
\begin{tabular}{lll|rrrr|l|rrrr|}
\cline{4-7} \cline{9-12}
 &  &  & \multicolumn{4}{l|}{\textbf{Answer accuracy (\%)}} &  & \multicolumn{4}{l|}{\textbf{Reference Accuracy (\%)}} \\ \cline{1-2} \cline{4-7} \cline{9-12} 
\multicolumn{1}{|l|}{\textbf{Dataset name}} & \multicolumn{1}{l|}{\textbf{Language}} &  & \multicolumn{1}{l|}{\textbf{Base}} & \multicolumn{1}{l|}{\textbf{\begin{tabular}[c]{@{}l@{}}All\\ Steps\end{tabular}}} & \multicolumn{1}{l|}{\textbf{\begin{tabular}[c]{@{}l@{}}w/o\\ Step 1\end{tabular}}} & \multicolumn{1}{l|}{\textbf{\begin{tabular}[c]{@{}l@{}}w/o\\ Step 3\end{tabular}}} &  & \multicolumn{1}{l|}{\textbf{Base}} & \multicolumn{1}{l|}{\textbf{\begin{tabular}[c]{@{}l@{}}All\\ Steps\end{tabular}}} & \multicolumn{1}{l|}{\textbf{\begin{tabular}[c]{@{}l@{}}w/o\\ Step 1\end{tabular}}} & \multicolumn{1}{l|}{\textbf{\begin{tabular}[c]{@{}l@{}}w/o\\ Step 3\end{tabular}}} \\ \cline{1-2} \cline{4-7} \cline{9-12} 
\multicolumn{1}{|l|}{\multirow{9}{*}{TyDi}} & \multicolumn{1}{l|}{Arabic} &  & \multicolumn{1}{r|}{89.7} & \multicolumn{1}{r|}{90.1} & \multicolumn{1}{r|}{89.6} & \textbf{91.2} &  & \multicolumn{1}{r|}{88.5} & \multicolumn{1}{r|}{93.5} & \multicolumn{1}{r|}{94.0} & \textbf{95.2} \\ \cline{2-2} \cline{4-7} \cline{9-12} 
\multicolumn{1}{|l|}{} & \multicolumn{1}{l|}{Bengali} &  & \multicolumn{1}{r|}{81.4} & \multicolumn{1}{r|}{\textbf{85.0}} & \multicolumn{1}{r|}{84.1} & \textbf{85.0} &  & \multicolumn{1}{r|}{74.3} & \multicolumn{1}{r|}{84.1} & \multicolumn{1}{r|}{\textbf{88.5}} & 86.7 \\ \cline{2-2} \cline{4-7} \cline{9-12} 
\multicolumn{1}{|l|}{} & \multicolumn{1}{l|}{English} &  & \multicolumn{1}{r|}{\textbf{87.5}} & \multicolumn{1}{r|}{85.9} & \multicolumn{1}{r|}{84.7} & 85.2 &  & \multicolumn{1}{r|}{91.4} & \multicolumn{1}{r|}{96.8} & \multicolumn{1}{r|}{96.3} & \textbf{97.3} \\ \cline{2-2} \cline{4-7} \cline{9-12} 
\multicolumn{1}{|l|}{} & \multicolumn{1}{l|}{Finnish} &  & \multicolumn{1}{r|}{81.2} & \multicolumn{1}{r|}{82.1} & \multicolumn{1}{r|}{81.6} & \textbf{83.8} &  & \multicolumn{1}{r|}{79.4} & \multicolumn{1}{r|}{87.7} & \multicolumn{1}{r|}{90.8} & \textbf{91.4} \\ \cline{2-2} \cline{4-7} \cline{9-12} 
\multicolumn{1}{|l|}{} & \multicolumn{1}{l|}{Indonesian} &  & \multicolumn{1}{r|}{88.1} & \multicolumn{1}{r|}{87.7} & \multicolumn{1}{r|}{88.0} & \textbf{90.0} &  & \multicolumn{1}{r|}{87.9} & \multicolumn{1}{r|}{94.3} & \multicolumn{1}{r|}{95.2} & \textbf{95.9} \\ \cline{2-2} \cline{4-7} \cline{9-12} 
\multicolumn{1}{|l|}{} & \multicolumn{1}{l|}{Korean} &  & \multicolumn{1}{r|}{87.0} & \multicolumn{1}{r|}{\textbf{88.0}} & \multicolumn{1}{r|}{84.8} & 85.5 &  & \multicolumn{1}{r|}{79.7} & \multicolumn{1}{r|}{88.4} & \multicolumn{1}{r|}{\textbf{89.9}} & 88.8 \\ \cline{2-2} \cline{4-7} \cline{9-12} 
\multicolumn{1}{|l|}{} & \multicolumn{1}{l|}{Russian} &  & \multicolumn{1}{r|}{86.6} & \multicolumn{1}{r|}{85.6} & \multicolumn{1}{r|}{85.8} & \textbf{86.9} &  & \multicolumn{1}{r|}{88.7} & \multicolumn{1}{r|}{89.9} & \multicolumn{1}{r|}{93.5} & \textbf{93.8} \\ \cline{2-2} \cline{4-7} \cline{9-12} 
\multicolumn{1}{|l|}{} & \multicolumn{1}{l|}{Swahili} &  & \multicolumn{1}{r|}{70.7} & \multicolumn{1}{r|}{72.7} & \multicolumn{1}{r|}{71.7} & \textbf{73.9} &  & \multicolumn{1}{r|}{79.8} & \multicolumn{1}{r|}{80.6} & \multicolumn{1}{r|}{80.2} & \textbf{85.2} \\ \cline{2-2} \cline{4-7} \cline{9-12} 
\multicolumn{1}{|l|}{} & \multicolumn{1}{l|}{Telugu} &  & \multicolumn{1}{r|}{79.2} & \multicolumn{1}{r|}{81.8} & \multicolumn{1}{r|}{\textbf{84.7}} & 82.6 &  & \multicolumn{1}{r|}{82.3} & \multicolumn{1}{r|}{90.8} & \multicolumn{1}{r|}{\textbf{95.0}} & 94.6 \\ \cline{1-2} \cline{4-7} \cline{9-12} 
\multicolumn{1}{|l|}{\multirow{12}{*}{XQuAD}} & \multicolumn{1}{l|}{Arabic} &  & \multicolumn{1}{r|}{77.4} & \multicolumn{1}{r|}{\textbf{80.6}} & \multicolumn{1}{r|}{77.2} & 77.9 &  & \multicolumn{1}{r|}{75.6} & \multicolumn{1}{r|}{90.6} & \multicolumn{1}{r|}{90.4} & \textbf{92.2} \\ \cline{2-2} \cline{4-7} \cline{9-12} 
\multicolumn{1}{|l|}{} & \multicolumn{1}{l|}{Chinese} &  & \multicolumn{1}{r|}{87.1} & \multicolumn{1}{r|}{\textbf{89.2}} & \multicolumn{1}{r|}{87.6} & 88.3 &  & \multicolumn{1}{r|}{82.4} & \multicolumn{1}{r|}{90.1} & \multicolumn{1}{r|}{92.3} & \textbf{94.0} \\ \cline{2-2} \cline{4-7} \cline{9-12} 
\multicolumn{1}{|l|}{} & \multicolumn{1}{l|}{English} &  & \multicolumn{1}{r|}{89.1} & \multicolumn{1}{r|}{\textbf{91.3}} & \multicolumn{1}{r|}{90.1} & 90.9 &  & \multicolumn{1}{r|}{79.2} & \multicolumn{1}{r|}{90.1} & \multicolumn{1}{r|}{92.4} & \textbf{94.2} \\ \cline{2-2} \cline{4-7} \cline{9-12} 
\multicolumn{1}{|l|}{} & \multicolumn{1}{l|}{German} &  & \multicolumn{1}{r|}{86.8} & \multicolumn{1}{r|}{88.0} & \multicolumn{1}{r|}{85.9} & \textbf{88.7} &  & \multicolumn{1}{r|}{75.4} & \multicolumn{1}{r|}{88.9} & \multicolumn{1}{r|}{91.0} & \textbf{92.6} \\ \cline{2-2} \cline{4-7} \cline{9-12} 
\multicolumn{1}{|l|}{} & \multicolumn{1}{l|}{Greek} &  & \multicolumn{1}{r|}{73.4} & \multicolumn{1}{r|}{\textbf{76.5}} & \multicolumn{1}{r|}{75.5} & 76.4 &  & \multicolumn{1}{r|}{57.4} & \multicolumn{1}{r|}{82.0} & \multicolumn{1}{r|}{84.6} & \textbf{89.5} \\ \cline{2-2} \cline{4-7} \cline{9-12} 
\multicolumn{1}{|l|}{} & \multicolumn{1}{l|}{Hindi} &  & \multicolumn{1}{r|}{70.0} & \multicolumn{1}{r|}{72.5} & \multicolumn{1}{r|}{71.0} & \textbf{72.8} &  & \multicolumn{1}{r|}{64.9} & \multicolumn{1}{r|}{83.5} & \multicolumn{1}{r|}{84.1} & \textbf{88.3} \\ \cline{2-2} \cline{4-7} \cline{9-12} 
\multicolumn{1}{|l|}{} & \multicolumn{1}{l|}{Romanian} &  & \multicolumn{1}{r|}{81.9} & \multicolumn{1}{r|}{\textbf{84.6}} & \multicolumn{1}{r|}{81.8} & 83.9 &  & \multicolumn{1}{r|}{72.6} & \multicolumn{1}{r|}{88.6} & \multicolumn{1}{r|}{89.9} & \textbf{92.1} \\ \cline{2-2} \cline{4-7} \cline{9-12} 
\multicolumn{1}{|l|}{} & \multicolumn{1}{l|}{Russian} &  & \multicolumn{1}{r|}{84.5} & \multicolumn{1}{r|}{86.3} & \multicolumn{1}{r|}{85.8} & \textbf{87.2} &  & \multicolumn{1}{r|}{76.7} & \multicolumn{1}{r|}{89.4} & \multicolumn{1}{r|}{89.8} & \textbf{92.9} \\ \cline{2-2} \cline{4-7} \cline{9-12} 
\multicolumn{1}{|l|}{} & \multicolumn{1}{l|}{Spanish} &  & \multicolumn{1}{r|}{85.9} & \multicolumn{1}{r|}{87.9} & \multicolumn{1}{r|}{86.6} & \textbf{88.1} &  & \multicolumn{1}{r|}{79.0} & \multicolumn{1}{r|}{90.6} & \multicolumn{1}{r|}{90.4} & \textbf{92.9} \\ \cline{2-2} \cline{4-7} \cline{9-12} 
\multicolumn{1}{|l|}{} & \multicolumn{1}{l|}{Thai} &  & \multicolumn{1}{r|}{78.4} & \multicolumn{1}{r|}{80.5} & \multicolumn{1}{r|}{80.2} & \textbf{80.9} &  & \multicolumn{1}{r|}{69.6} & \multicolumn{1}{r|}{85.9} & \multicolumn{1}{r|}{90.4} & \textbf{91.3} \\ \cline{2-2} \cline{4-7} \cline{9-12} 
\multicolumn{1}{|l|}{} & \multicolumn{1}{l|}{Turkish} &  & \multicolumn{1}{r|}{75.5} & \multicolumn{1}{r|}{\textbf{75.8}} & \multicolumn{1}{r|}{72.0} & 74.3 &  & \multicolumn{1}{r|}{71.1} & \multicolumn{1}{r|}{86.8} & \multicolumn{1}{r|}{89.2} & \textbf{92.0} \\ \cline{2-2} \cline{4-7} \cline{9-12} 
\multicolumn{1}{|l|}{} & \multicolumn{1}{l|}{Vietnamese} &  & \multicolumn{1}{r|}{83.0} & \multicolumn{1}{r|}{84.6} & \multicolumn{1}{r|}{82.5} & \textbf{85.8} &  & \multicolumn{1}{r|}{74.3} & \multicolumn{1}{r|}{88.8} & \multicolumn{1}{r|}{91.2} & \textbf{93.1} \\ \cline{1-2} \cline{4-7} \cline{9-12} 
\end{tabular}
\caption{(Continued) Full per language answer and reference accuracies for each dataset.}
\label{tab:fullresultsappendix2}
\end{table*}